# ISPO: An Integrated Ontology of Symptom Phenotypes for Semantic Integration of Traditional Chinese Medical Data

Zixin Shu[a#], Rui Hua[a#], Dengying Yan[a], Chenxia Lu[b], Ning Xu[c], Jun Li[d], Hui Zhu[d], Jia Zhang[b], Dan Zhao[e], Chenyang Hui[f], Junqiu Ye[f], Chu Liao[d], Qi Hao[d], Wen Ye[d], Cheng Luo[d], Xinyan Wang[a], Chuang Cheng[a], Xiaodong Li[b], Baoyan Liu[c], Xiaji Zhou[g], Runshun Zhang[g], Min Xu[h], Xuezhong Zhou[a*]

[a]Institute of Medical Intelligence, School of Computer and Information Technology, Beijing Jiaotong University, Beijing, 100063, China

[b]Hubei Key Laboratory of the theory and application research of liver and kidney in traditional Chinese medicine, Hubei Provincial Hospital of Traditional Chinese Medicine, Wuhan, 430061, China

[c]National Data Center of Traditional Chinese Medicine, China Academy of Chinese Medical Sciences, Beijing, 100700, China

[d]Clinical College of Traditional Chinese Medicine, Hubei University of Chinese Medicine, Wuhan, 430000, China

[e]Department of Laboratory Medicine, Hubei Provincial Hospital of Traditional Chinese Medicine, Wuhan, 430061, China

[f]Department of Infectious Diseases, Hubei Provincial Hospital of Traditional Chinese Medicine, Wuhan, 430061, China

[g]Guang'anmen Hospital, China Academy of Chinese Medical Sciences, Beijing, 100053, China

[h]Information Technology Center, the First Affiliated Hospital, College of Medicine, Zhejiang University, Hangzhou, 310003, China

[#]Authors contributed equally

Corresponding Authors

*Prof Xuezhong ZHOU PhD

Institute of Medical Intelligence, School of Computer and Information Technology, Xizhimenwai Street, Haidian District, Jiaotong University, Beijing 100063, China.

Email: xzzhou@bjtu.edu.cn

# ABSTRACT

**Background:** Symptom phenotypes are one of the key types of manifestations for diagnosis and treatment of various disease conditions. However, the diversity of symptom terminologies is one of the major obstacles hindering the analysis and knowledge sharing of various types of symptom-related medical data particularly in the fields of Traditional Chinese Medicine (TCM). Objective: This study aimed to construct an Integrated Ontology of symptom phenotypes (ISPO) to support the data mining of Chinese EMRs and real-world study in TCM field.

**Methods:** To construct an integrated ontology of symptom phenotypes (ISPO), we manually annotated classical TCM textbooks and large-scale Chinese electronic medical records (EMRs) to collect symptom terms with support from a medical text annotation system. Furthermore, to facilitate the semantic interoperability between different terminologies, we incorporated public available biomedical vocabularies by manual mapping between Chinese terms and English terms with cross-references to source vocabularies. In addition, we evaluated the ISPO using independent clinical EMRs to provide a high-usable medical ontology for clinical data analysis.

**Results:** By integrating 78,696 inpatient cases of EMRs, 5 biomedical vocabularies, 21 TCM books and dictionaries, ISPO provides 3,147 concepts, 23,475 terms, and 55,552 definition or contextual texts. Adhering to the taxonomical structure of the related anatomical systems of symptom phenotypes, ISPO provides 12 top-level categories and 79 middle-level sub-categories. The validation of data analysis showed the ISPO has a coverage rate of 95.35%, 98.53% and 92.66% for symptom terms with occurrence rates of 0.5% in additional three independent curated clinical datasets, which can demonstrate the significant value of ISPO in mapping clinical terms to ontologies.

**Conclusions:** ISPO delivers an integrated controlled vocabulary for symptom phenotypes that would enhance semantic interoperability among heterogeneous medical data sources and clinical decision support systems in TCM fields.

# 1. Introduction

Symptom phenotypes (i.e., symptoms and signs), which could be obtained by human natural perception and cognition abilities, are one of the main clinical manifestations of disease conditions[1, 2]. Systematic investigation on symptom phenotypes could help discover the cause of disease, promote the development of drug, and improve clinical diagnosis and individualized treatment[3]. Traditional Chinese Medicine (TCM) is a kind of traditional medicine encompassing the entire range of clinical empirical knowledge on clinical manifestations and herbal therapies with thousands years of long history[4]. In the field of TCM, there are numerous Chinese symptom terms present in clinical EMRs[5]. Different organizations and research institutions face significant obstacles when it comes to mining knowledge related to symptom concepts for the diversity of expression of terms, resulting in a lack of comparability and consistency in research findings[6, 7]. Therefore, the standardization and conceptualization of symptom terminology are crucial for the real-world clinical research of TCM[8].

Ontology and controlled vocabularies propose a powerful approach to explicitly specify the medical concepts and their categories for supporting the data integration and semantic-based analysis of large-scale data in medical fields [9-11]. The famous projects of biomedical ontology and terminologies, such as Gene Ontology[12], Medical Subject Headings (MeSH)[13], Systematized Nomenclature of Medicine Clinical Terms (SNOMED CT)[14] and human disease ontology[15], proposed both high-quality phenotype and genotype terminological references for bioinformatics and biomedical informatics. In particular, the Unified Medical Language System (UMLS)[16] is an integrated medical language system developed by the US National Library of Medicine, which realizes the integration and dissemination of a series of terminological knowledge sources (including SNOMED CT). Biomedical Informatics Ontology System (BIOS), the first large-scale publicly available biomedical knowledge graph generated completely by machine learning algorithms, which contains 4.1 million concepts, 7.4 million terms and 7.3 million relation triplets[17]. In addition, to facilitate the sharing and use of various public medical terminological resources, BioPortal (http://bioportal.bioontology.org) proposed an open repository of biomedical ontologies[18], range in subjects including anatomy, phenotype, experimental conditions, imaging, chemistry and health. In TCM fields, the development of UTCMLS early time in the 2000s [19], an integrated medical language system similar to UMLS, has aimed to facilitate the TCM bibliographic literature semantic searching and associated data analysis. Related study developed a TCM-Grid that help health professionals, researchers, enterprises and personal users to retrieve, integrate and share TCM information and knowledge from geographically decentralized TCM database resources and knowledge base resources in China[20]. In recent years, TCM knowledge graph sources, such as SymMap[21] and SoFDA[22], were also developed to provide important terminological sources for network pharmacology studies.

Although there established symptom-related ontologies (e.g. Symptom Ontology (SO) [23] and Human Phenotype Ontology (HPO) [24]) and the symptom sub-categories in large-scale medical terminologies (e.g. MeSH[13]), which have played important roles for biomedical informatics research[25-28], it still has several essential issues that need be addressed to improve their actual utilities in TCM clinical data analysis. First, the range of terms in most symptom-related ontologies are still very limited. For example, SO is an ontology specifically designed for the description and classification of symptoms, providing definitions, synonyms, preferred terms, and a hierarchical structure for symptoms, but it is still difficult to cover common symptoms in EMRs. SO provides limited descriptions of the semantic content of common symptoms (i.e., a limited number of sub-nodes), failing to adequately reflect the clinical diversity of nonspecific symptoms such as headaches and vomiting, each of which has only one sub-node. Furthermore, SO contains 944 symptom concepts with 1,174 symptom terms, which has a limited number of established synonymous term relationships, thereby constraining its practical utility in clinical and research contexts. HPO mainly covers human phenotype terms of congenital genetic diseases and only obtains ~200 (1.27%) concepts after filtering symptoms in HPO by limiting the semantic types of concepts as Sign or Symptom (i.e., T184) in UMLS. Second, there are significant differences in the understanding the symptom and signs between TCM and modern medicine, primarily due to fundamentally different theoretical foundations, diagnostic methods, and treatment strategies. Modern medicine relies on the biomedical model, focusing on anatomy, physiology, and biochemical mechanisms. Consequently, symptoms are often confused with diseases, laboratory tests, and medical imaging in biomedical ontologies. TCM clinical symptoms mainly include the patient's subjective discomfort and the signs observed by traditional examination methods such as palpation, percussion, and auscultation. The observations of tongue and pulse characteristics by physicians are also important signs that need to be referred to in the process of individualized diagnosis and treatment. For example, the concept "Corneal Endothelial Cell Loss" (D055954), under the C23.888 sub-category (i.e., Symptom and Signs Category) in MeSH is actually a pathological manifestation or diagnosis within the context of TCM clinical research [32]. Furthermore, modern medicine focuses more on specific symptoms that may indicate the progression of a disease. However, TCM pays more attention to non-specific symptoms, such as diet, daily routines, bowel movements, sleep, and emotions, which emphasize the principle of prevention before the onset of disease. Finally, most available medical terminologies lack high-quality hierarchy structures to accurately reflect the semantic relationships between symptom concepts. For example, *Nasal Discharge (SYMP_0000701)* and *Yellow Exudate from Nose (SYMP_0000664)* are more closely related to one another than to *Nasal Bleeding (SYMP_0000741)*. However, these three concepts are placed under the nose sub-category of SO as sibling concepts, which does not strictly reflect the actual semantic closeness of these concepts.

Currently, there is still no project on large-scale integrated symptom ontology in Chinese to support the data mining of Chinese EMRs and real-world study in TCM field. In order to facilitate the data integration and semantic analysis of TCM clinical data and those curated data derived from biomedical literatures, we combined the joint efforts of 20 experts in many fields, including informatics, computer science and TCM to manually construct an integrated symptom phenotype ontology (ISPO) covering the utilization of terminologies from TCM. By integrating the symptom concepts from 30 symptom terminological sources including TCM clinical EMRs, biomedical controlled vocabularies, TCM books and dictionaries, ISPO provides 3,147 concepts, 23,475 terms, and 55,552 definition or contextual texts. ISPO delivers an integrated controlled vocabulary for symptom phenotypes, with both clinical and biomedical literature synonyms in Chinese and English languages. ISPO would enhance the semantic interoperability among heterogeneous medical data sources and clinical decision support systems, in particular for Chinese medical big data involving the features of symptom phenotypes.

## 2. Methods

**2.1 The design principle and developing framework of ISPO**

We developed ISPO that could be used for both TCM clinical data analysis and biomedical data mining on symptom phenotypes includes the following main purposes and applications:

1) Standardizing clinical symptom terminology in TCM to improve the quality and comparability of clinical data, which promotes the informatization and modernization construction of TCM clinical practice.

2) Enhancing the comparability of real-world clinical research in TCM, thereby improving the understanding of clinical diseases and assisting researchers in better managing and studying clinical symptoms.

3) Facilitating the sharing of symptom knowledge between traditional medicine and modern medicine, providing necessary data support for the study of symptom phenotypes and molecular mechanisms.

ISPO integrates synonyms and hierarchical relationships, which are fundamental semantic structures in ontology development. Synonym relationships within ISPO ensure that different terms conveying the same meaning are linked, enhancing the ontology's capacity to map equivalent terms across various datasets. Additionally, hierarchical relationships establish a structured taxonomy, organizing terms from general to specific and reflecting their conceptual relationships. The initial version of ISPO was based on high-frequency symptom terms in TCM EMRs. It expanded the terminology from classical TCM books and dictionaries, and mapped symptom concepts from biomedical ontologies such as the UMLS. Here, we utilized a human-machine collaborative approach for terminology editing and ontology construction. This process includes key steps such as terminology collection, concept mapping, hierarchy structure construction, and ontology visualization (Fig. 1).

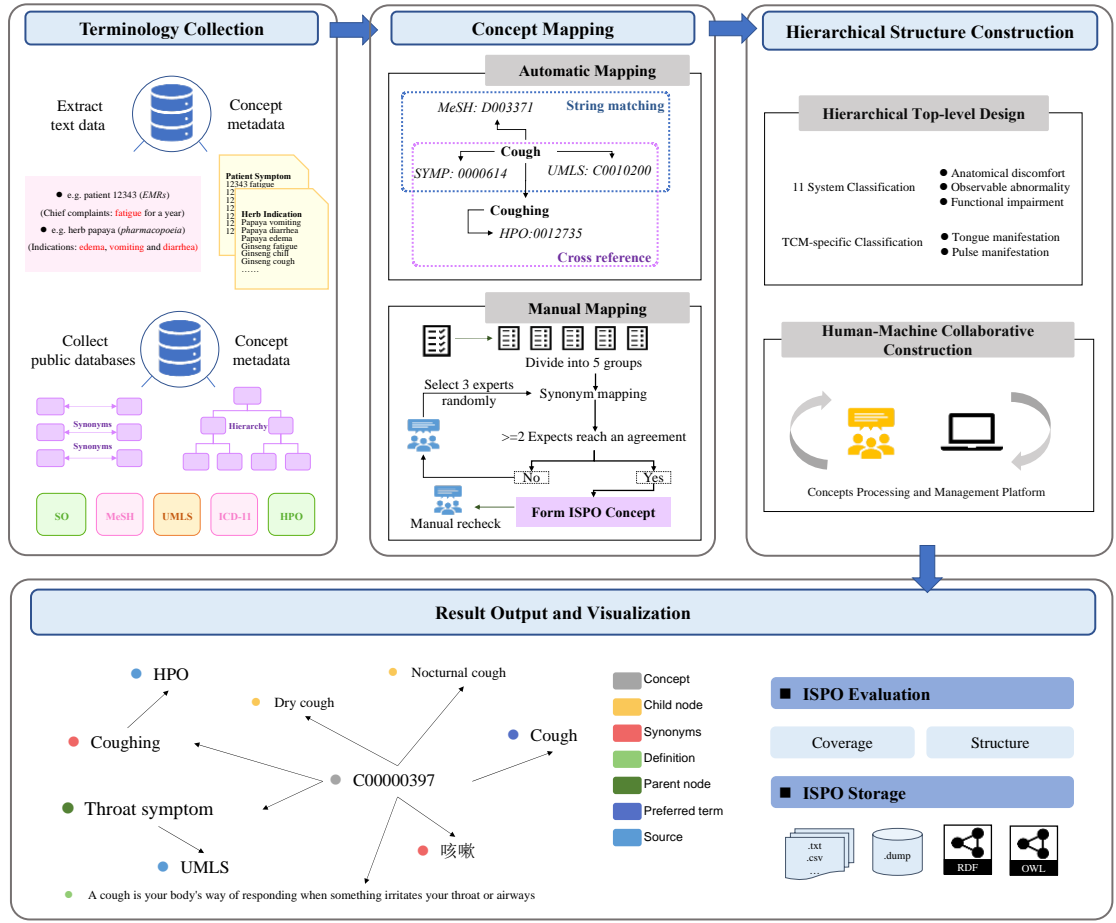

Fig. 1. Overview of the ISPO construction framework. The main processes of ISPO construction include terminology collection, concept mapping, hierarchy structure construction and ontology visualization. We annotated and extracted symptom entities from TCM texts using the clinical information extraction tools. Subsequently, we manually screened symptom concepts from the biomedical control vocabulary which were integrated based on the cross-reference relationship among these vocabularies. All symptom terms were reviewed and distributed to 15 domain experts for manual mapping, including the mapping of Chinese ancient and modern terms, as well as Chinese and English terms. Then, we constructed the ontology hierarchy using a bottom-up approach and assigned classification codes to each ISPO concept.

**2.2 Data sources and collection**

ISPO collected and integrated symptom terms from TCM clinical EMRs of 78,696 patients, 21 ancient and contemporary TCM textbooks and dictionaries, and 5 biomedical controlled vocabularies, including:

1) TCM clinical EMRs: To curate various types of symptom clinical terms from EMRs [29], we collected Chinese EMRs of 78,696 cases from 7 highest-level TCM hospitals, including seven hundreds of chronic diseases. We have shown the distribution of 30 diseases with over 500 cases, such as chronic lung disease, heart disease, stroke, cancer, hypertension and kidney disease (Fig. S1 in Supplementary File 1). We extracted symptom terms from admission records in large-scale EMRs based on a clinical information extraction tool (HCPSA, See Methods)[30], which facilitates the task of manual annotations of medical entities and collection of clinical symptom terms. Finally, we extracted and reviewed 3,963 symptom entities with more than one record from EMRs that were considered to be common symptoms.

2) TCM textbooks and dictionaries: To integrate the symptom terminologies in TCM books and dictionaries[31], we also manually annotated 14 ancient TCM books, 7 contemporary textbooks and dictionaries with support from HCPSA, and collected 5,379 symptom entities.

3) Biomedical controlled vocabularies: To increase semantic interoperability between biomedical literatures (in Egnlish) and clinical data (in Chinese), we incorporated public available biomedical controlled vocabularies including UMLS (Version 2020), MeSH (Version 2022), SO (Version 2022), ICD (Version 11th, ICD-11) and HPO (Version 2022). Then, we filter and collected 16,334 concepts under the category of symptoms and signs, including SO (944 concepts) and sub-categories of C23.888 in MeSH (399 concepts), Chapter 21 in ICD-11 (1,234 concepts), the T184 in UMLS (13,757 concepts) and HPO (226 concepts). Table S1 and S2 provide basic information on data sources of symptom terminologies.

**2.3 Symptom concept mapping**

Although concept mapping could be treated as a supervised learning task [32-34], to accumulate high-quality cross-references of concepts between different terminologies, manual concept mapping is vital to obtain high reliable results. Therefore, we invited 15 medical researchers to participate in the manual mapping task under the guidance of domain experts. First, clinical terms were randomly and equally divided into 5 groups, and terms in each group were randomly assigned to 3 experts for manual mapping. For the mapping results of the same term, it needs to be determined by the unanimous agreement of two or more experts. After the repeated steps of manual mapping and review, terms were finally mapped to unique ISPO concepts. In addition, by utilizing the currently available concept associations between existing biomedical controlled vocabularies, a small number of English concept mappings can be constructed based on cross-references of strings and codes, which mainly contribute to the expansion of English terminologies (Fig. 1).

In ISPO, we designed a data structure keeping the information of cross-references to source vocabularies that refers to those of the UMLS to integrate the established terminologies and their codes for both clinical settings and biomedical literature. Each record is connected by a Concept Unique Identifier (CUI), a String Unique Identifier (synonyms for the concept, SUI), and an Atom Unique Identifier (AUI), which organizes multiple different terms of a concept in an orderly manner. For example, we collected both English and Chinese terms of the cough symptom from MeSH, UMLS, and Chinese books, then manually mapped these terms with cross-languages to form the unified concept of cough with unified identifier (i.e.C00000397) in ISPO (Fig. 2).

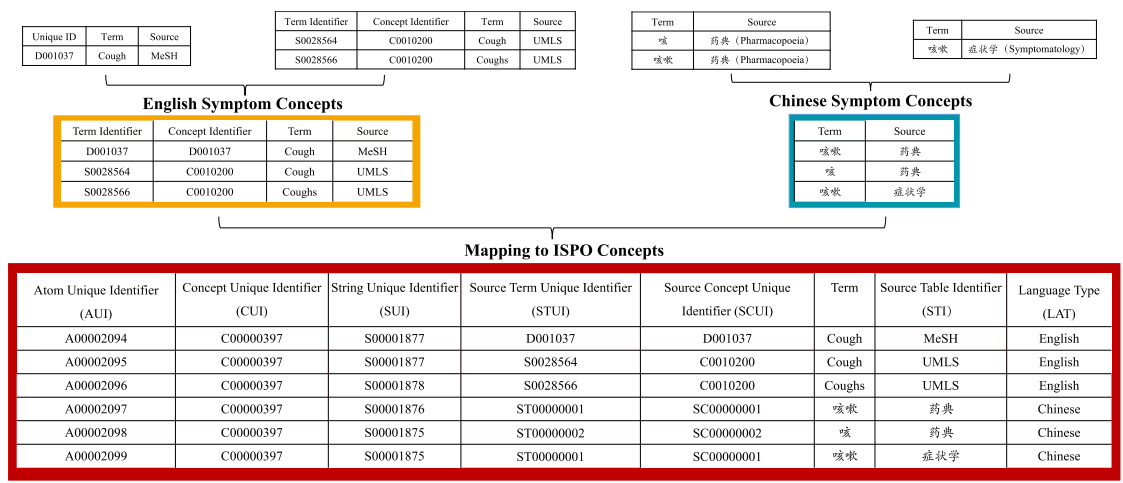

Fig. 2 Cross-language synonym mapping of cough terms.

## 2.4 Concept taxonomy construction

ISPO is constructed in a bottom-up manner by 15 medical researchers using a multi-user collaborative terminology processing and management system (MCSP-PS, See Methods). In ISPO, each concept has only one parent concept within the hierarchy tree, and the classification process of each concept requires review by experts. Referring to the top-level structure of established biomedical ontologies (e.g. MeSH) and the taxonomical structure of anatomical systems related to symptom phenotypes, we designed ISPO with 11 anatomical system categories and one TCM tongue and pulse signs category as top-level categories (Table 1).

Table 1. ISPO top-level categories are designed according to the biomedical vocabulary.

| ISPO Top-level categories | MeSH (C23.888 categories) | SO Top-level categories | ICD-11 (Chapter 21 categories) |
|---|---|---|---|
| Nervous system symptoms | Neurologic Manifestations | Nervous system symptom | Symptoms or signs involving the nervous system |
| Respiratory system symptoms | Signs and Symptoms, Respiratory | Respiratory system and chest symptom | Symptoms or signs involving the respiratory system |
| Circulatory system symptoms | None | Cardiovascular system symptom | Symptoms or signs involving the circulatory system |
| Digestive system symptoms | Signs and Symptoms, Digestive | Digestive system symptom | Symptoms or signs involving the digestive system or abdomen |
| Musculoskeletal system symptoms | *None* | Musculoskeletal system symptom | Symptoms or signs of the musculoskeletal system |
| Urinary system symptoms | Urological Manifestations | Urinary system symptom | Symptoms, signs or clinical findings of the genitourinary system |
| Mental and behavioral symptoms | *None* | Neurological and physiological symptom | Mental or behavioural symptoms, signs or clinical findings |
| Skin and integumentary system symptoms | Skin Manifestations | skin and integumentary tissue symptom | Symptom or signs involving the skin |
| Reproductive system symptoms | *None* | Reproductive system symptom | *None* |
| General symptoms | Medically Unexplained Symptoms | General symptom | General symptoms |
| Nutrition, metabolism, and development symptoms | *None* | Nutrition, metabolism, and development symptom | Symptoms of endocrine, nutritional or metabolic diseases |
| TCM tongue and pulse signs | *None* | *None* | *None* |

Compared to established biomedical ontologies, the ISPO extends to include categories of tongue and pulse signs that are unique to TCM clinics. Tongue manifestation includes descriptions of tongue quality, fur and sublingual vessels. Pulse manifestation records the characteristics of the pulse as felt under the fingers by a physician, including location, rhythm and strength (Table 2). The tongue and pulse signs recorded in TCM clinical EMRs are critically important for diagnosing and treating diseases. For example, TCM recognized the red tongue with yellow fur coating typically indicates a heat syndrome, and when it co-occurs with a wiry pulse, it is often associated with damp-heat syndrome in the liver and gallbladder. This condition is also often accompanied by symptoms such as fever, a bitter taste in the mouth, and rib pain.

Table 2. The category of tongue and pulse in ISPO.

| Category | Definition |
|---|---|
| Tongue manifestation | |
| Tongue quality | Tongue quality, also referred to as the "tongue body", relates to the musculature of the tongue. |
| Tongue color | The colors of the tongue quality include shades such as pale white, red, purple, and blue tongues. |

| Tongue shape | The shapes of the tongue quality such as tough, tender, thin, plump and other specific pathological forms (e.g., prickly tongue). |
|---|---|
| Tongue condition | The dynamics of the tongue body include flexible and free movement, which are considered normal. Pathological conditions of the tongue include flaccidity, stiffness, and trembling. |
| Tongue fur | Tongue fur refers to a moss-like layer on the surface of the tongue that indicates various health conditions. |
| Fur character | The characteristics of the tongue fur include thickness, moisture (e.g., moist or dry), texture (e.g., smooth or rough) and distribution (e.g., partial or complete). |
| Fur color | The colors of the tongue fur include white, yellow, gray, black and green. |
| Sublingual vessel | The sublingual vessels are the thicker blue-purple veins on either side beneath the tongue. |
| Pulse manifestation | |
| Floating pulse class | Floating pulses are characterized by their shallow positioning and easy accessibility upon palpation, including types such as floating, surging, and scattered pulses. |
| Deep pulse class | Deep pulses are characterized by their deeper positioning, only detectable with firm pressure, and include types such as deep, deep-sited, and weak pulses. |
| Slow pulse class | Slow pulses are characterized by a slower pulse rate, typically less than four to five beats per breath (equivalent to less than 60 beats per minute), and include types such as slow, moderate, and hesitant pulses. |
| Rapid pulse class | Rapid pulses are characterized by a faster pulse rate, exceeding five beats per breath (equivalent to more than 90 beats per minute), and include types such as rapid, swift, and irregular-rapid pulses. |
| Feeble pulse class | Feeble pulses referred to as "deficient pulses", are characterized by a feeling of feebleness upon palpation at all three positions (cun, guan, and chi), representing an overall lack of strength in the pulse. |
| Excess pulse class | Excess pulses are characterized by strong pulses felt at all three positions (cun, guan, and chi), and include types such as slippery, stringy and tight pulses. |
| Unusual pulse class | There are ten abnormal pulse types observed when life is at risk, including the oblique flying pulse and ectopic radial pulse. |

**2.5 ISPO construction tools**

To ensure efficiency and quality in the development of ISPO, our team has developed two systems designed to assist with the construction of ontologies, including Human-machine Cooperative Phenotypic Spectrum Annotation System (HCPSAS, www.tcmai.org) [30], and Medical Concept Structure and Relationship Processing System (MCSR-PS, http://www.tcmkg.com/ISPO/MCSR-PS).

HCPSAS facilitates structured annotation tasks on a variety of unstructured texts, including entities, relationships and events. HCPSAS integrates both a rule-based knowledge acquisition algorithm and a deep learning model architecture featuring BiLSTM-CRF. The platform also provides a professional review system to monitor and ensure the quality of the annotated data. We utilize HCPSAS to annotate various entities from TCM texts, such as positive symptoms, negative symptoms, and tongue and pulse signs (Fig. 3). Following the annotation, users submit their results to the database and conduct lexicon verification on the annotated entities. Through the intelligent iteration interface, users can filter entities by type and frequency in the lexicon and then use the verified entities to efficiently pre-annotate all samples, thereby achieving continuous optimization and precise annotation work.

**Admission record:**

**Chief complaint**: insomnia accompanied by palpitations for over three months.

**Current symptoms observed**: poor sleep quality, difficulty falling asleep, early waking, repeated awakening after falling asleep, occasional palpitations, daytime drowsiness, hot in hands and feet, dry mouth and bitter taste, frequent sighing, loose stools, hot urine, dark yellow urine, no frequent urination, no symptoms of dizziness, headache, chest tightness, shortness of breath, nausea or vomiting and other discomfort, poor spirits, reduced appetite, red tongue with yellow coating, wiry and rapid pulse.

Legend: Negative Symptoms, Positive Symptoms, Tongue and Pulse, Location, Frequency

⬇ Annotation Results

| # | Entity Text | Start Position | End Position | Tag Category | Result Type | Review Status |
|---|---|---|---|---|---|---|
| 1 | insomnia | 4 | 6 | Positive Symptoms | Iteration | Approved |
| 2 | palpitations | 7 | 9 | Positive Symptoms | Iteration | Approved |
| 3 | poor sleep quality | 18 | 20 | Positive Symptoms | Iteration | Approved |
| 4 | difficulty falling asleep | 21 | 24 | Positive Symptoms | Manual | Approved |
| …… | occasional palpitations | 35 | 39 | Positive Symptoms | Manual | Approved |
| …… | occasional | 35 | 37 | Frequency | Manual | Approved |

Fig. 3. TCM clinical EMRs annotation page of the HCPSAS.

MCSR-PS is a multi-user collaborative terminology processing and management system that using the Python programming language to assist with ontology hierarchy construction. It can import file in OWL, RDF, and other formats for terminology processing and supports operations such as inserting, deleting, and automatically coding terms, which significantly reduces the time required for terminology processing in our tasks. As shown in Fig. 4, after uploading the ontology file, users can access the MCSR-PS editing page. The left side of the interface displays the ontology tree structure, where users can drag, edit, delete, and add nodes at the concept level. On the right side, users can manage synonymous terms by adding, deleting, or modifying them using the corresponding concept code. The system also supports both global and local term searches and allows for the editing and deleting of entities during searches. Logging operations at the bottom of the page enhances team collaboration efficiency and the system supports multi-user collaborative editing.

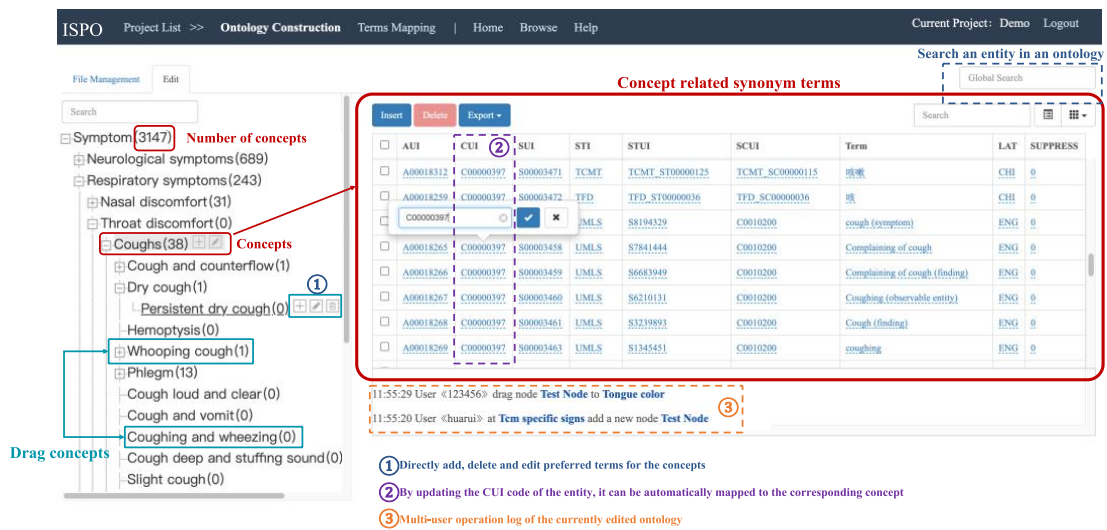

Fig. 4. Interface of MCSR-PS for term editing.

## 3. Results

ISPO is proposed as a public available web resource (http://www.tcmkg.com/ISPO/home) for free use and has been uploaded to BioPortal (https://bioportal.bioontology.org/ontologies/ISPO/) in May 2023.

### 3.1 Category structure of ISPO

Adhering to the taxonomical structure of the related anatomical systems of symptom phenotypes, ISPO provides 12 top-level categories. Among these, nervous system symptoms (689 concepts, 21.89%), digestive system symptoms (545 concepts, 17.32%), and musculoskeletal system symptoms (492 concepts, 15.63%) rank as the top three when sorted by concept capacity (Fig. 5A). We compared the categories of source concepts and found that ~20% of concepts in ancient TCM books were classified as general symptoms, whereas less than 4% of concepts in biomedical vocabularies, suggesting that traditional medicine may focus more on describing non-specific symptoms than modern medicine (Fig. S2 in Supplementary File 1). Furthermore, we analyzed the distribution characteristics of SO concepts across top-level categories since ISPO and SO have a similar anatomical system classification structure. 832 SO concepts have category definitions (Fig. 5A), including nervous system symptoms (149 concepts,17.91%) and the head and neck symptoms (102 concepts,12.26%). In addition, we integrated 476 SO concepts into 388 ISPO concepts through cross-language mapping and compared the top-level category features of these concepts in the two ontologies to explore their similarities and differences in taxonomic structures. The results showed that most concepts had similar top-level categories, such as those for respiratory system symptoms and urinary system symptoms (Fig. 5B).

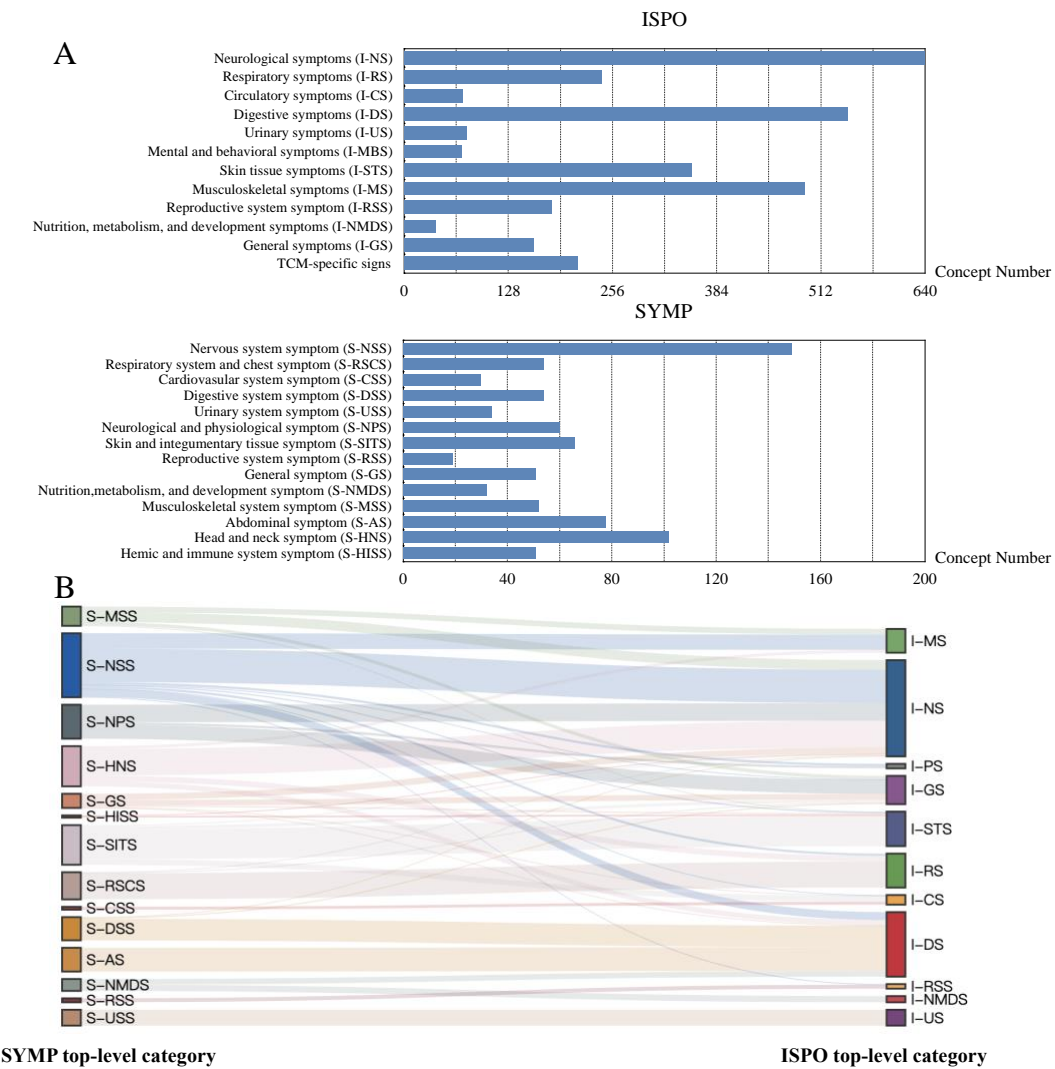

Fig. 5. The comparison of symptom concepts in ISPO and SO. A) Concept distribution of the top-level categories in SO and ISPO, B) The mapping relationship between SO and ISPO system category. Here, 476 concepts from SO were mapped into 388 ISPO concepts. We compared the top-level categories of these concepts in SO and ISPO. For example, fatigue belongs to the category of neurological and physiological symptoms in SO, while it belongs to the category of general symptoms in ISPO.

**3.2 Symptom concepts in ISPO**

We formed ISPO concepts through manual mapping from various source vocabularies. Initially, we extracted 3,963 terms from TCM EMRs and 5,379 terms from TCM literature, mapping them to form 3,147 ISPO concepts. Furthermore, we integrated English symptom vocabularies that includes 16,334 concepts from biomedical vocabularies. Through Chinese-English mapping, we mapped 3,192 of these English concepts to ISPO concepts and added 12,770 English synonyms from biomedical vocabularies. However, since TCM symptom terminology often utilizes the classical Chinese language, known for its conciseness and expressiveness, there are still 1,229 ISPO concepts lacking English synonym terms. For example, "消谷善饥" (Xiao Gu Shan Ji) depicts a condition of rapid digestion and overeating, while "心中憺憺大动" (Xin Zhong Dan Dan Da Dong) describes severe palpitations accompanied by intense panic and anxiety. Therefore, we used manual translation and expert review to expand the English synonyms for these ISPO concepts. Finally, we obtained 3,147 unique ISPO concepts and 23,475 terms.

Here, ISPO selects the high-frequency terms as preferred terms for symptom concepts according to clinical data analysis[35]. Furthermore, considering that concepts can express different connotations in specific contexts [51], ISPO also provides 55,552 contextual texts for symptoms, including 4,495 definition texts from public available resources (e.g. Wikipedia), 3,105 reference texts from ancient TCM books, and 47,952 chief complaint records from EMRs. These contextual texts can provide the characteristics, attributes, and relationships of concepts, which can help the ontology reasoning engine better understand the relationships and semantics between concepts[36, 37].

### 3.3 Structural and coverage evaluation of ISPO

Ontology evaluation is designed to measure the quality of an ontology, either to provide feedback to ontology developers and knowledge engineers or to provide their users with insights on the adequacy of the ontology[38]. Here, we evaluated the reliability of the ontology with structural features[39] and calculated the coverage of clinical terms to assess the current capacity of concepts in ISPO[40].

The structural features of an ontology include topological and logical properties that are measured by means of context-free metrics (i.e., depth and breadth, related to the cardinality of paths in a graph) [47, 48]. We compared the tree structures of ISPO and SO based on core structural features (Table 3). The results showed that ISPO had a deeper structure with an average depth of leaf nodes at 4.76, and also had a broader scope with an average breadth of 314.7, indicating a larger number of classes across various levels.

Table 3. The structural parameters measured for ISPO and SO

| Features | Root count | Class count | Synonym count | Leave node count | Maximum Depth | Average depth | Average width |
|---|---|---|---|---|---|---|---|
| ISPO | 12 | 3,147 | 23, 475 | 2,279 | 10 | 4.76 | 314.7 |
| SO | 14 | 889 | 1135 | 712 | 7 | 3.14 | 127.0 |

Since ISPO was constructed to facilitate data integration and semantic analysis of mainly TCM clinical data, we manually annotated symptom terms in additional TCM EMRs through HCPSAS to evaluate clinical symptom coverage, including admission records of 40,800 patients with liver cirrhosis (HBTCMC) and 12,626 patients with steatohepatitis (HBTCMS) from Hubei Provincial Hospital of TCM. Furthermore, we also extracted symptom terms from the EMRs of 48,057 inpatients across all departments of Shanxi Provincial Hospital of TCM (SXTCM). We observed a long-tail distribution of disease and symptom entities within our datasets, with many conditions affecting only a small number of patients. This phenomenon not only indicates the high incidence of certain clinical conditions but also the diversity in expressions across numerous clinical terms. We have presented the distribution of 50 prevalent diseases, each with over 500 cases in the datasets, which exhibit substantial variability in their distribution. For example, liver cirrhosis and hepatitis B are more common in the HBTCMC and HBTCMS datasets, while hypertension and type 2 diabetes have a more consistent presence across all datasets, indicating their widespread prevalence in clinical settings (Fig. 6A).

Based on the obtained symptom terms and their corresponding entities counts (i.e., patient numbers), we conducted a screening process to identify clinical terms with an occurrence rate of at least 0.01% in the sample population and reviewed the accuracy of the annotations. The results showed that ISPO demonstrated high entity coverage rates in three datasets: 93.29% for the HBTCMC dataset, 97.26% for the HBTCMS dataset and 94.98% for the SXTCM dataset (Table 4).

Table 4. Evaluation coverage rate of ISPO on clinical symptoms.

| Dataset | Sample size | Entity number | Term number | Entities covered | Terms covered |
|---|---|---|---|---|---|
| HBTCMC | 40,800 | 168,709 | 648 | 157,392 (93.29%) | 595 (91.98%) |
| HBTCMS | 12,626 | 74,630 | 804 | 72,582 (97.26%) | 619 (76.99%) |
| SXTCM | 48,057 | 278,433 | 1,132 | 264,461 (94.98%) | 780 (68.90%) |

Furthermore, we categorized these clinical terms according to their occurrence rates in the sample population and evaluated the coverage effectiveness of ISPO in covering various levels of the occurrence frequencies. The results showed that ISPO can achieve impressive coverage rates: 95.35% for HBTCMC, 98.53% for HBTCMS, and 92.66% for SXTCM, for symptom terms with an occurrence rate of at least 0.5% in the medical records (Fig. 6B). These findings highlight the effectiveness of ISPO in capturing common symptom terms and its highest coverage for nervous system symptoms and digestive system symptoms (Fig. 6C).

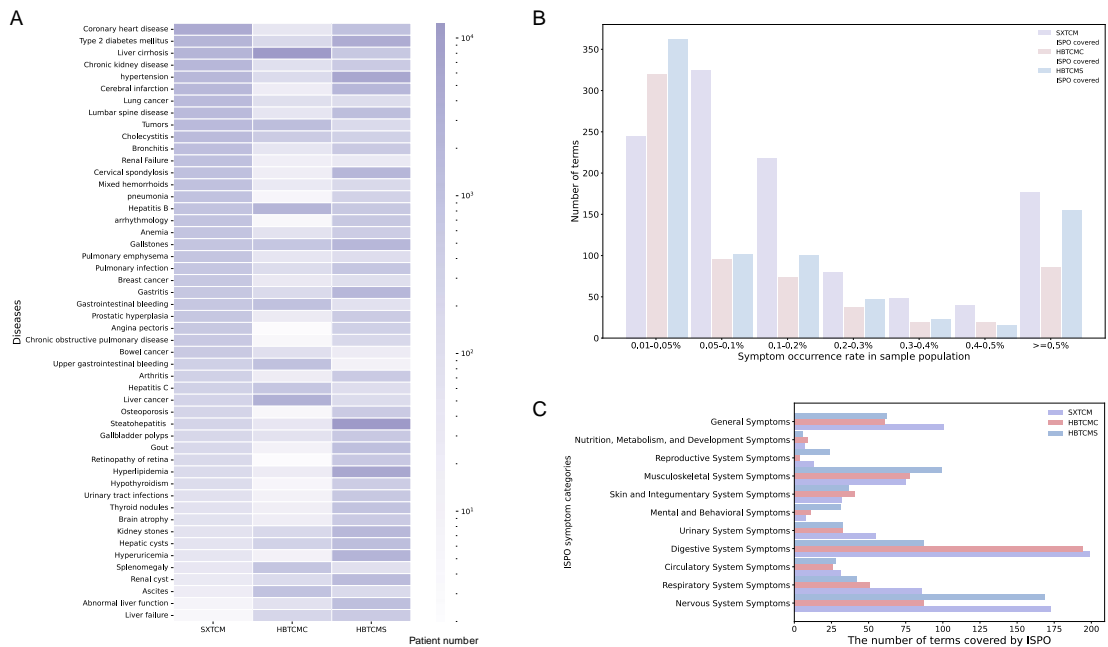

Fig. 6. Overview of term distribution in clinical datasets. A) Distribution of 50 common diseases within three datasets; B) Coverage distribution of symptom terms by ISPO in datasets; C) Distribution of categories for symptom terms covered by ISPO in datasets.

Furthermore, we categorized these symptom terms according to their occurrence rates in the sample population and evaluated the coverage effectiveness of ISPO in covering various levels of occurrence frequencies. The results showed that ISPO achieved 61.40%, 50.42% and 17.83% coverage of symptom concepts in MeSH, SO, and ICD-11, respectively. Actually, the low overlap rate of these three terminology sources is mainly own to the fact that many symptom terms currently included in these sources are actually not symptom phenotypes, but belong to diseases, laboratory tests, syndromes, pathology or physiology. After manual review of these term types by medical terminological experts (Table 5), we found that SO included 203 (21.50%) diseases, 46 (9.02%) symptom categories, and 16 (1.69%) laboratory tests. In particular, many symptom terms in the ICD-11 contain the descriptions of "unspecified" or "other specified", such as the concept *Other specified ascites (ICD-11_ME04.Y)*, which lacks specific meaning and therefore is difficult to be mapped to ISPO. By manually re-examining these concept types, we found that ISPO actually can reach about 81.38%, 74.16% and 49.40% coverage of symptom concepts in MeSH, SO and ICD-11.

Table 5. The type distributions of concepts in ICD-11, SO and MeSH

| Concept types | SO | MeSH (C23.888) | ICD-11 Chapter 21 | Examples |
|---|---|---|---|---|
| Symptom | 627 (66.42%) | 290 (72.68%) | 417 (33.79%) | Fever (MeSH_D005334), Cough (MeSH_D003371), Coma (MeSH_D003128), Chest pain (MeSH_D002637), Fatigue (SYMP_0019177) |
| Symptom Category | 46 (4.87%) | 13 (3.26%) | 62 (5.02%) | Abdominal symptom (SYMP_0000461), General symptom (SYMP_0000567), Lower urinary tract symptoms (MeSH_D059411), Urological manifestations (MeSH_D020924), Symptom or complaint of the knee (ICD-11_ME86.A) |
| Syndrome | 5 (0.53%) | 22 (5.51%) | 13 (1.05%) | Gerstmann syndrome (MeSH_D005862), Kearns-sayre syndrome (MeSH_D007625), Korsakoff syndrome (MeSH_D020915), Horner syndrome (MeSH_D006732), Shoulder syndrome (ICD-11_ME86.D0) |
| Disease | 203 (21.50%) | 32 (8.02%) | 98 (7.94%) | Arthritis (SYMP_0019169), Rhinitis (SYMP_0000121), Stroke (SYMP_0000734), Pneumonia (SYMP_0019168), Hepatic abscess (SYMP_0000295) |
| Pathology/ Physiology | 34 (3.60%) | 21 (5.26%) | 60 (4.86%) | Hypoventilation (MeSH_D007040), Blood vessel infection (SYMP_0000227), Lesions in myocardium (SYMP_0000070), Vein swelling (SYMP_0000168), Pulmonary consolidation (SYMP_0000729) |
| Laboratory Test | 16 (1.69%) | 9 (2.26%) | 186 (15.07%) | Cytopenia (SYMP_0000691), Leukopenia (SYMP_0000317), Albuminuria (MeSH_D000419), Positive occult blood in stool (ICD-11_ME24.A6), Elevated creactive protein (ICD-11_MA14.15) |
| Other Description | 13 (1.38%) | 12 (3.01%) | 398 (32.25%) | Other specified ascites (ICD-11_ME04.Y), Other specified chronic pain (ICD-11_MG30.Y), Unattended death (ICD-11_MH13), Birth weight (MeSH_D001724), Weight cycling (MeSH_D000091622) |

## 4. Applications

### 4.1 Ontology browser and semantic search

The ISPO web platform (http://www.tcmkg.com/ISPO/home) helps users to navigate symptom concepts and browse semantic information about a concept, such as synonyms, hypernyms, semantic types, definitions, comments, and associated concepts. It enables users to navigate through the conceptual system in the following two ways: 1) Navigate through the ISPO concept hierarchy. The browser uses a tree-view to present the hyponymy relations between symptom concepts. When a user clicks a concept in the tree-view, the browser presents the detailed semantic information about the concept in the main panel. 2) Navigate through the ISPO semantic network. when a user selects a concept, the browser presents a graphical view of the semantic network around the concept.

Moreover, this web system utilizes the content of the ontology to implement functions such as synonymous search, providing more comprehensive and accurate results for users. When a user searches for a term, the system retrieves all the synonyms of the term and uses them to return extended search results. The system also considers semantic relations in query processing. For example, if the user enters the search term "cough", the system will find the definition text of cough, ancient literature, and annotation information.

**4.2 Entity linking for clinical symptom terms**

Mapping clinical terms to symptom ontologies for normalization and standardization is one of the key applications of clinical ontologies in EMR data. Through this approach, individualized data can be read, understood, and analyzed by machines, thus enabling the classification, aggregation, and statistical analysis of symptom data to better assist medical personnel in clinical data analysis and improving their understanding of disease diagnosis and treatment[41, 42]. However, systematic mapping of the large number of terms contained in clinical data such as EMRs, to a standard ontology requires intensive manual efforts with domain expertise. Therefore, we have developed a framework to better assist in mapping clinical terminology to ISPO. In this framework, we consider the task of entity linking as a sequence-to-sequence learning[43] problem and employ two LSTM[44] networks as the encoder and decoder. Training is conducted on a manually constructed benchmark dataset consisting of approximately 5,000 mapping rules. For example, "head and facial skin pain" can be mapped into "headache" and "facial skin pain", which would be transformed to a rule instance using "head and facial skin pain" as the input sequence and "headache/facial skin pain" as the output sequence. The dataset was divided into a training set and a test set in a ratio of 8:2 and it can achieve 84.6% accuracy on the test set after training. We have distributed this model in ISPO web platform that offers a useful tool for users who need clinical terminology standardization of their symptom list. Furthermore, to facilitate the practical mapping of a given clinical symptom list or vocabulary, we integrated three sequential steps incorporating the Seq2Seq model to help map the clinical terms to ISPO efficiently and accurately. First, we will use exact mapping step to map terms through exact string matching directly to ISPO concept codes. Next, we would use automatic mapping to make predictions on source terms and generate multiple predicted ISPO concepts as candidates for manual evaluation. Finally, we would support the task of rule mapping for manual efforts to curated mapping rules on the remainder terms or compound terms to generate multiple ISPO concept codes.

**4.3 ISPO in TCM real-world clinical research**

To address the heterogeneity of symptoms in real-world clinical data, we employed the ISPO, which significantly enhanced the consistency and relatedness of symptom terms, providing a foundation for data analysis in various TCM clinical studies. In previous study, we explored the effects of TCM treatments on COVID-19 in 1,788 patients across five hospitals in Wuhan[45]. Using the HCPSA system, we annotated 3,188 positive symptom entities from EMRs and encompassing 111 symptom terms. ISPO enabled us to unify diverse symptom expressions, such as "fatigue", "overall weakness", and "tiredness" into a unified concept. Moreover, ISPO allows for the effective clustering of otherwise sparse data through hierarchical organization. For example, although "throat discomfort" was only recorded once, ISPO helped identify 24 patients with related symptoms, including sore throat, itchy throat, and dry throat (Table 6). Finally, ISPO standardized 111 terms into 52 symptom concepts, reducing the dimensions by 53%. Based on standardized the clinical data, we used propensity score matching method to demonstrate for the first time that TCM prescription treatment can significantly reduce the mortality and improved the symptoms of COVID-19 patients. This case demonstrates the crucial role of ISPO in effectively integrating and standardizing TCM clinical data.

Table 6. Impact of ISPO on clinical symptom term standardization

| Original symptom | Expanded symptom terms via ISPO[1] | Patient count (pre-process)[2] | Patient count (post-process)[3] |
|---|---|---|---|
| Fever | Fever, Intermittent fever, Recurrent fever, Low-grade fever, Subjective fever, Intermittent low-grade fever | 1,130 | 1,276 |
| Cough | Cough, Dry cough, Cough with phlegm, Occasional cough, Recurrent cough, Intermittent cough | 906 | 1,000 |
| Dyspnea | Shortness of breath, Wheezing, Breathlessness, Labored breathing, Dyspnea, Post-activity dyspnea, Intermittent breathlessness | 36 | 183 |
| Fatigue | Fatigue, Limb weakness, Intermittent fatigue, Limb fatigue, Tiredness, Overall weakness | 139 | 156 |
| Chest tightness | Chest tightness, Feeling of suffocation, Intermittent chest tightness, Chest constriction discomfort | 143 | 148 |
| Difficulty breathing | Difficulty breathing, Labored breathing | 64 | 65 |
| Diarrhea | Diarrhea, Mild diarrhea | 47 | 48 |
| Poor appetite | Poor appetite, Decreased appetite, Anorexia, Reduced appetite, Lackluster appetite | 22 | 34 |
| Throat discomfort | Throat discomfort, Throat pain, Dry throat, Throat itchiness, Throat redness and itchiness, Itchy throat | 1 | 24 |
| Body pain | Body pain, Body ache, Body soreness, Limb pain | 8 | 20 |

[1]Expanded terms via ISPO includes the original symptom term along with any subordinate or synonymous terms that have been unified under this symptom category by ISPO. [2]Patient count (pre-processing) represents the number of patients documented with the original symptom term before applying ISPO. This count reflects the initial recording of symptoms as they were originally noted in clinical data. [3]Patient count (post-processing) shows the actual number of patients suffering from the symptom after processing through ISPO.

**4.4 ISPO in clinical and biomolecular integration research**

ISPO has facilitated mapping between TCM clinical terminology and biomedical vocabularies, advancing the integration of clinical and biomolecular research. In a recent clinical study, we assessed the effectiveness of herb-symptom associations in 1,936 liver cirrhosis patients[46]. Utilizing ISPO, 648 clinical symptoms extracted from EMRs were standardized into 269 unified concepts, achieving a dimension reduction of 58.49%. Furthermore, to integrate clinical patient data with genetic data, we also utilized ISPO to create mappings between Chinese symptom terms from clinical data and their corresponding English terms from the UMLS. Consequently, 315 English symptom terms associated with genetic data mapped to 92 Chinese symptom terms in the patient data. For example, several UMLS codes, such as C4042891 (sleep-wake disorders) and C0917801 (sleep disorder insomnia), mapped to the Chinese term 失眠 (insomnia).

Based on the related symptom knowledge, we established a network medicine framework and revealed the topological relationships between herb-symptom associations within the human protein interactome. Moreover, we validated the effectiveness of this framework in predicting the proximity between herbs and symptoms using patient data, thereby demonstrating the significant value of ISPO in facilitating the integration of clinical and biomolecular research.

## 4. Discussion

Ontology development has been an active research topic to propose semantic solution for integrating various data sources in biomedical field[38, 41, 42, 47]. Various developed biomedical ontologies have facilitated knowledge management and querying, data integration, knowledge inference, and reasoning for clinical decision support and precision medicine [5, 12, 15, 16, 24]. With growing interests in the effectiveness and safety of TCM, an increasing number of research institutions and scholars have performed real-world clinical studies to assess the practical therapies of TCM[48-50]. These studies utilize diverse data sources, such as electronic health records, to uncover the overall diagnosis patterns and treatment regularities of TCM in real-world clinical settings.

However, a key challenge in TCM clinical big data research lies in the lack of clinical ontological resources, particularly those related to symptom terminology[51]. In fact, typical symptom terminologies are rarely developed with large concept capacity, which might largely be due to the fact that most symptom information is recorded in free text format in EMRs, and structured entry of symptom phenotypes in clinical settings is labor-intensive [52]. Currently, the tasks of standardized symptom terminologies in real-world TCM studies rely on manual efforts by individual researchers, lacking a comprehensive symptom ontology repository to facilitate semantic interoperability and ensure the comparability of research findings[6, 53, 54]. To ensure the standardization and universality of symptom entities in EMRs in TCM clinical practice, since 2018, we have organized more than 20 related researchers to manually collect symptom terms from various data sources. Although the construction process takes a significant amount of time and manual efforts, ISPO has been primarily constructed manually by researchers to ensure the quality of the ontology.

By unifying the descriptions and terminologies of symptoms, ISPO proposes a large-scale symptom ontology with 3,147 concepts and 23,475 terms. The symptom terms provided by ISPO would largely fulfill the needs of real-world clinical research in TCM, with the coverage rates for high-frequency terms reaching 95.35% (HBTCMC), 98.53% (HBTCMS) and 92.66% (SXTCM) in three independent curated clinical datasets. ISPO have provided critical support for term mapping and association in our previous various clinical studies. For example, ISPO standardized a large number of clinical symptom terms to assist in disease (e.g., liver disease[48]) subgroup classification research based on clinical features. Furthermore, ISPO provides 8,943 Chinese terms and 14,532 English terms for 3,147 symptom concepts, which would be useful for the integration of symptom-related data resources in both Chinese and English languages, thereby advancing the field of symptom science[55].

Due to the diagnosis requirements and medical reimbursement for clinical patients, medical ontologies related to disease phenotypes have been intensively developed with an increasing number of concepts and codes, which would often include those terms of symptom phenotypes as sub-categories [17, 62, 63]. For example, a sub-category with the tree code of C23.888 in MeSH incorporates the concepts of symptoms and signs mainly used in the context of biomedical literatures [14], which contains totally 399 concepts. However, when manually evaluated the semantic types of those terms, we found that 8.02% of them are actually diseases, 5.51% are syndromes, and 5.26% are pathology or physiology terms. In addition, HPO [25] mainly contains the phenotype terms of human genetic disease conditions aims to facilitate the clinical differentiation and detection of the biological mechanisms of human genetic diseases. The 17,512 phenotype concepts in HPO cover various aspects of human phenotype features, including organ defects, diseases, symptoms, and metabolic abnormalities. However,

HPO only contains 226 symptom concepts when specifically filtered according to the semantic types of concepts as T184 in UMLS. Although CHPO (the Chinese version of HPO) has been developed, the number of symptom terms is actually limited, and it only covers 329 concepts (10.45%) in ISPO. SO is typical symptom terminology to standardize the concepts and definitions of both symptoms and signs, with around 1,000 concepts in 2022 version[28]. However, the similar condition of the mixing types of phenotypes still exists for SO and we found that about 34% of the SO concepts are not symptoms (Table 2).

Our work has several limitations. Firstly, the capacity of symptom concepts is mainly limited to high-frequently used unitary clinical symptom terms (e.g., fever, headache and cough). The compound symptom terms with combinations of unitary clinical terms (e.g., severe foot pain in joints, mainly the big toe) were not included. In the future, we will attempt to associate compound symptoms with unitary symptoms as semantic relationships by incorporating their manifestation attributes such as anatomical locations, degree, nature and temporal aspects, to help represent more specific symptom phenotypes. Secondly, considering that symptoms often involve multiple systems and etiologies, the taxonomical structure of the symptom phenotypes still needs to be continuously optimized with possible multiple hierarchies as those of MeSH [13]. Finally, to fully exert the biomedical potential of ISPO, it would be necessary and promising to develop a knowledge graph containing various relationships with symptoms(e.g. symptom-gene relationships[56], symptom-disease relationships[57] and drug-symptom relationships[21]), thereby further improving the usability and semantic integration capability of ISPO[58].

Nevertheless, the fully development of ISPO would have the potential not only to promote curation of high-quality clinical data[9] for real-world TCM clinical research[59], but also to help advance the symptom science[2] since a high-quality symptom ontology is one of the vital tasks that should be developed to integrate the various symptom related data sources in medical field.

## 5. Conclusions

ISPO delivers an integrated controlled vocabulary for symptom phenotypes, with both clinical and biomedical literature synonyms in Chinese and English languages, which aims to enhance the semantic operationality and comparability of data analysis in TCM fields.

## Acknowledgments

This work is partially supported by the National Natural Science Foundation of China (82174533 and 82204941), the Natural Science Foundation of Beijing (M21012) and the Key project of Hubei Natural Science Foundation (2020CFA023).

## Authors' contributions

XZ Zhou, M Xu, RS Zhang, XJ Zhou, XD Li and BY Liu conceived the study. ZX Shu and R Hua analyzed the data. XY Wang, C Cheng and R Hua developed the ontology editing system. All authors provided important contributions to data collection, processing and review. ZX Shu, R Hua and XZ Zhou drafted and revised the manuscript. All authors have proofread the manuscript.

## Conflicts of Interest

None declared.

# References

1. A. K. Cashion, J. Gill, R. Hawes, W. A. Henderson, L. Saligan, National Institutes of Health Symptom Science Model sheds light on patient symptoms, Nurs Outlook. 64 (5) (2016) 499-506. https://doi.org/10.1016/j.outlook.2016.05.008.
2. M. Dodd, S. Janson, N. Facione, J. Faucett, E. S. Froelicher, J. Humphreys, K. Lee, C. Miaskowski, K. Puntillo, S. Rankin, Advancing the science of symptom management, J Adv Nurs. 33 (5) (2001) 668-676. https://doi.org/10.1046/j.1365-2648.2001.01697.x.
3. K. Hickey, S. Bakken, M. Byrne, D. Bailey, G. Demiris, S. Docherty, S. Dorsey, B. Guthrie, M. Heitkemper, C. Jacelon, Corrigendum to precision health: advancing symptom and self-management science, Nurs Outlook. 68 (2) (2020) 139-140. https://doi.org/10.1016/j.outlook.2019.11.003.
4. X. Zhou, Z. Wu, A. Yin, L. Wu, W. Fan, R. Zhang, Ontology development for unified traditional Chinese medical language system, Artif. Intell. Med. 32 (1) (2004) 15-27. https://doi.org/10.1016/j.artmed.2004.01.014.
5. L. Li, P. Wang, J. Yan, Y. Wang, S. Li, J. Jiang, Z. Sun, B. Tang, T.-H. Chang, S. Wang, Real-world data medical knowledge graph: construction and applications, Artif. Intell. Med. 103 (2020) 101817. https://doi.org/10.1016/j.artmed.2020.101817.
6. W. Sun, Z. Cai, Y. Li, F. Liu, S. Fang, G. Wang, Data processing and text mining technologies on electronic medical records: a review, J. Healthcare Eng. 2018 (2018). https://doi.org/10.1155/2018/4302425.
7. W. Raghupathi, V. Raghupathi, Big data analytics in healthcare: promise and potential, Health information science and systems. 2 (2014) 1-10. https://doi.org/10.1186/2047-2501-2-3.
8. X. Zhou, Y. Li, Y. Peng, J. Hu, R. Zhang, L. He, Y. Wang, L. Jiang, S. Yan, P. Li, Clinical phenotype network: the underlying mechanism for personalized diagnosis and treatment of traditional Chinese medicine, Frontiers of medicine. 8 (2014) 337-346. https://doi.org/10.1007/s11684-014-0349-8.
9. R. Hoehndorf, P. N. Schofield, G. V. Gkoutos, The role of ontologies in biological and biomedical research: a functional perspective, Briefings Bioinf. 16 (6) (2015) 1069-1080. https://doi.org/10.1093/bib/bbv011.
10. T. R. Gruber, A translation approach to portable ontology specifications, Knowl Acquis. 5 (2) (1993) 199-220. https://doi.org/10.1006/knac.1993.1008.
11. A. L. Rector, W. A. Nowlan, G. Consortium, The GALEN project, Comput. Methods Programs Biomed. 45 (1-2) (1994) 75-8. https://doi.org/10.1016/0169-2607(94)90020-5.
12. M. Ashburner, C. A. Ball, J. A. Blake, D. Botstein, H. Butler, J. M. Cherry, A. P. Davis, K. Dolinski, S. S. Dwight, J. T. Eppig, Gene ontology: tool for the unification of biology, Nat. Genet. 25 (1) (2000) 25-9. https://doi.org/10.1038/75556.
13. C. E. Lipscomb, Medical subject headings (MeSH), Bulletin of the Medical Library Association. 88 (3) (2000) 265.
14. M. Q. Stearns, C. Price, K. A. Spackman, A. Y. Wang. SNOMED clinical terms: overview of the development process and project status. in *Proceedings of the AMIA Symposium*. 2001. American Medical Informatics Association

15. L. M. Schriml, C. Arze, S. Nadendla, Y.-W. W. Chang, M. Mazaitis, V. Felix, G. Feng, W. A. Kibbe, Disease Ontology: a backbone for disease semantic integration, Nucleic Acids Res. 40 (D1) (2012) D940-D946. https://doi.org/10.1093/nar/gkr972.
16. O. Bodenreider, The unified medical language system (UMLS): integrating biomedical terminology, Nucleic Acids Res. 32 (suppl_1) (2004) D267-D270. https://doi.org/10.1093/nar/gkh061.
17. S. Yu, Z. Yuan, J. Xia, S. Luo, H. Ying, S. Zeng, J. Ren, H. Yuan, Z. Zhao, Y. Lin, Bios: An algorithmically generated biomedical knowledge graph, arXiv preprint arXiv:2203.09975. (2022).
18. M. A. Musen, N. F. Noy, N. H. Shah, P. L. Whetzel, C. G. Chute, M.-A. Story, B. Smith, a. t. N. team, The National Center for Biomedical Ontology, J. Am. Med. Inf. Assoc. 19 (2) (2011) 190-195. https://doi.org/10.1136/amiajnl-2011-000523.
19. H. Long, Y. Zhu, L. Jia, B. Gao, J. Liu, L. Liu, H. Herre, An ontological framework for the formalization, organization and usage of TCM-Knowledge, BMC Med. Inf. Decis. Making. 19 (2) (2019) 79-89. https://doi.org/10.1186/s12911-019-0760-9.
20. H. Chen, Z. Wu, C. Huang, J. Xu. TCM-Grid: weaving a medical grid for traditional Chinese medicine. in *Computational Science—ICCS 2003: International Conference, Melbourne, Australia and St. Petersburg, Russia June 2–4, 2003 Proceedings, Part III 3*. 2003. Springer
21. Y. Wu, F. Zhang, K. Yang, S. Fang, D. Bu, H. Li, L. Sun, H. Hu, K. Gao, W. Wang, SymMap: an integrative database of traditional Chinese medicine enhanced by symptom mapping, Nucleic Acids Res. 47 (D1) (2019) D1110-D1117. https://doi.org/10.1093/nar/gky1021.
22. Y. Zhang, N. Wang, X. Du, T. Chen, Z. Yu, Y. Qin, W. Chen, M. Yu, P. Wang, H. Zhang, SoFDA: an integrated web platform from syndrome ontology to network-based evaluation of disease-syndrome-formula associations for precision medicine, Sci. Bull. 67 (11) (2022) 1097-1101. https://doi.org/10.1016/j.scib.2022.03.013.
23. N. F. Noy, N. H. Shah, P. L. Whetzel, B. Dai, M. Dorf, N. Griffith, C. Jonquet, D. L. Rubin, M.-A. Storey, C. G. Chute, BioPortal: ontologies and integrated data resources at the click of a mouse, Nucleic Acids Res. 37 (suppl_2) (2009) W170-W173. https://doi.org/10.1093/nar/gkp440.
24. P. N. Robinson, S. Köhler, S. Bauer, D. Seelow, D. Horn, S. Mundlos, The Human Phenotype Ontology: a tool for annotating and analyzing human hereditary disease, Am. J. Hum. Genet. 83 (5) (2008) 610-615. https://doi.org/10.1016/j.ajhg.2008.09.017.
25. O. Mohammed, R. Benlamri, S. Fong. Building a diseases symptoms ontology for medical diagnosis: an integrative approach. in *The First International Conference on Future Generation Communication Technologies*. 2012. IEEE
26. E. Xia, W. Sun, J. Mei, E. Xu, K. Wang, Y. Qin. Mining disease-symptom relation from massive biomedical literature and its application in severe disease diagnosis. in *AMIA Annu. Symp. Proc.* 2018. American Medical Informatics Association
27. S. Dhiman, A. Thukral, P. Bedi. OHF: An Ontology Based Framework for Healthcare. in *International Conference on Artificial Intelligence and Speech Technology*. 2021. Springer.https://doi.org/10.1007/978-3-030-95711-7_28.
28. The Symptom Ontology (SYMP) [Online]. Available: https://bioportal.bioontology.org/ontologies/SYMP, (2022).

29. R. C. Wasserman, Electronic medical records (EMRs), epidemiology, and epistemology: reflections on EMRs and future pediatric clinical research, Academic pediatrics. 11 (4) (2011) 280-287. https://doi.org/10.1016/j.acap.2011.02.007.
30. Q. Zou, K. Yang, K. Chang, X. Zhang, X. Li, X. Zhou. Phenonizer: A fine-grained phenotypic named entity recognizer for Chinese clinical texts. in *2021 IEEE International Conference on Bioinformatics and Biomedicine (BIBM)*. 2021. IEEE.https://doi.org/10.1155/2022/3524090.
31. X. Zhou, Y. Peng, B. Liu, Text mining for traditional Chinese medical knowledge discovery: a survey, J. Biomed. Inf. 43 (4) (2010) 650-660. https://doi.org/10.1016/j.jbi.2010.01.002.
32. F. Dhombres, O. Bodenreider, Interoperability between phenotypes in research and healthcare terminologies—Investigating partial mappings between HPO and SNOMED CT, J. Biomed. Semant. 7 (2016) 1-13.
33. R. Hoehndorf, A. Oellrich, D. Rebholz-Schuhmann, Interoperability between phenotype and anatomy ontologies, Bioinformatics. 26 (24) (2010) 3112-3118. https://doi.org/10.1093/bioinformatics/btq578.
34. T. J. Callahan, A. L. Stefanski, J. M. Wyrwa, C. Zeng, A. Ostropolets, J. M. Banda, W. A. Baumgartner Jr, R. D. Boyce, E. Casiraghi, B. D. Coleman, Ontologizing health systems data at scale: making translational discovery a reality, arXiv preprint arXiv:2209.04732. (2022). https://doi.org/10.48550/arXiv.2209.04732.
35. N. Guarino, P. Giaretta, Ontologies and knowledge bases, Towards very large knowledge bases. (1995) 1-2.
36. S. Montani, How to use contextual knowledge in medical case-based reasoning systems: A survey on very recent trends, Artif. Intell. Med. 51 (2) (2011) 125-131. https://doi.org/10.1016/j.artmed.2010.09.004.
37. D. Benslimane, A. Arara, G. Falquet, Z. Maamar, P. Thiran, F. Gargouri. Contextual ontologies: Motivations, challenges, and solutions. in *Advances in Information Systems: 4th International Conference, ADVIS 2006, Izmir, Turkey, October 18-20, 2006. Proceedings 4*. 2006. Springer
38. H. Hlomani, D. Stacey, Approaches, methods, metrics, measures, and subjectivity in ontology evaluation: A survey, Semantic Web Journal. 1 (5) (2014) 1-11.
39. A. Gómez-Pérez, Ontology evaluation, in *Handbook on ontologies*. 2004, Springer. p. 251-273.
40. J. Brank, M. Grobelnik, D. Mladenic. A survey of ontology evaluation techniques. in *Proceedings of the conference on data mining and data warehouses (SiKDD 2005)*. 2005. Citeseer Ljubljana Slovenia
41. I. Harrow, R. Balakrishnan, E. Jimenez-Ruiz, S. Jupp, J. Lomax, J. Reed, M. Romacker, C. Senger, A. Splendiani, J. Wilson, Ontology mapping for semantically enabled applications, Drug Discovery Today. 24 (10) (2019) 2068-2075. https://doi.org/10.1016/j.drudis.2019.05.020.
42. M. Ivanović, Z. Budimac, An overview of ontologies and data resources in medical domains, Expert Syst. Appl. 41 (11) (2014) 5158-5166. https://doi.org/10.1016/j.eswa.2014.02.045.
43. I. Sutskever, O. Vinyals, Q. V. Le, Sequence to sequence learning with neural networks, Advances in neural information processing systems. 27 (2014).
44. S. Hochreiter, J. Schmidhuber, Long short-term memory, Neural Comput. 9 (8) (1997) 1735-1780. https://doi.org/10.1162/neco.1997.9.8.1735.

45. K. W. Chan, Z. Shu, K. Chang, B. Liu, X. Zhou, X. Li, Add-on Chinese Medicine for Coronavirus Disease 2019 (COVID-19): A Retrospective Cohort, European Journal of Integrative Medicine. 48 (2021) 101903.
46. X. Gan, Z. Shu, X. Wang, D. Yan, J. Li, S. Ofaim, R. Albert, X. Li, B. Liu, X. Zhou, Network medicine framework reveals generic herb-symptom effectiveness of traditional Chinese medicine, Science advances. 9 (43) (2023) eadh0215.
47. S. Cheng, X. Liang, Z. Bi, N. Zhang, H. Chen, ProteinKG65: a knowledge graph for protein science, arXiv preprint arXiv:2207.10080. (2022).
48. Z. Shu, W. Liu, H. Wu, M. Xiao, D. Wu, T. Cao, M. Ren, J. Tao, C. Zhang, T. He, Symptom-based network classification identifies distinct clinical subgroups of liver diseases with common molecular pathways, Computer methods and programs in biomedicine. 174 (2019) 41-50.
49. Z. Shu, Y. Zhou, K. Chang, J. Liu, X. Min, Q. Zhang, J. Sun, Y. Xiong, Q. Zou, Q. Zheng, Clinical features and the traditional Chinese medicine therapeutic characteristics of 293 COVID-19 inpatient cases, Frontiers of medicine. 14 (2020) 760-775. https://doi.org/10.1007/s11684-020-0803-8.
50. N. Xu, K. Zhong, H. Yu, Z. Shu, K. Chang, Q. Zheng, H. Tian, L. Zhou, W. Wang, Y. Qu, Add-on Chinese medicine for hospitalized chronic obstructive pulmonary disease (CHOP): A cohort study of hospital registry, Phytomedicine. 109 (2023) 154586.
51. B. Liu, X. Zhou, Y. Wang, J. Hu, L. He, R. Zhang, S. Chen, Y. Guo, Data processing and analysis in real-world traditional Chinese medicine clinical data: challenges and approaches, Stat. Med. 31 (7) (2012) 653-660. https://doi.org/10.1002/sim.4417.
52. Y. An, X. Xia, X. Chen, F.-X. Wu, J. Wang, Chinese clinical named entity recognition via multi-head self-attention based BiLSTM-CRF, Artif. Intell. Med. 127 (2022) 102282. https://doi.org/10.1016/j.artmed.2022.102282.
53. Z.-y. Gao, H. Xu, D.-z. Shi, C. Wen, B.-y. Liu, Analysis on outcome of 5284 patients with coronary artery disease: the role of integrative medicine, J. Ethnopharmacol. 141 (2) (2012) 578-583. https://doi.org/10.1016/j.jep.2011.08.071.
54. Z. Mou, L. He, Q. Zheng, Q. Cheng, Z. Shu, J. Yang, X. Zhou, B. Liu, Classification and Analysis of Symptom Characteristics and Acupoint Experience in Acupuncture Treatment of Children with Cerebral Palsy, World Science and Technology Modernization of Traditional Chinese Medicine 22 (11) (2020) 3959-3965 (in Chinese).
55. E. J. Corwin, J. A. Berg, T. S. Armstrong, A. D. Dabbs, K. A. Lee, P. Meek, N. Redeker, Envisioning the future in symptom science, Nurs Outlook. 62 (5) (2014) 346-351. https://doi.org/10.1016/j.outlook.2014.06.006.
56. K. Yang, N. Wang, G. Liu, R. Wang, J. Yu, R. Zhang, J. Chen, X. Zhou, Heterogeneous network embedding for identifying symptom candidate genes, J. Am. Med. Inf. Assoc. 25 (11) (2018) 1452-1459. https://doi.org/10.1093/jamia/ocy117.
57. X. Zhou, J. Menche, A.-L. Barabási, A. Sharma, Human symptoms–disease network, Nat. Commun. 5 (1) (2014) 4212. https://doi.org/10.1038/ncomms5212.
58. D. N. Nicholson, C. S. Greene, Constructing knowledge graphs and their biomedical applications, Comput. Struct. Biotechnol. J. 18 (2020) 1414-1428. https://doi.org/10.1016/j.csbj.2020.05.017.

59. X. Zhou, B. Liu, X. Zhang, Q. Xie, R. Zhang, Y. Wang, Y. Peng, eds. *Data mining in real-world traditional Chinese medicine clinical data warehouse*. Data Analytics for Traditional Chinese Medicine Research. 2013, Springer, Cham. 189-213.

# Supplementary File

## Supplementary Table S1. The basic information on data sources of symptom terminologies.

| Index | Data Source | Abbreviation | Basic Information |
|---|---|---|---|
| **Electronic medical records (EMRs)** | | | |
| 1 | Shandong provincial hospital of TCM | SDTC (hypertension) | EMRs of patients with hypertension in Shandong provincial hospital of TCM |
| 2 | Shanxi Provincial Hospital of TCM | SXTC | EMRs of patients in Shanxi Provincial Hospital of TCM |
| 3 | Hubei provincial hospital of TCM | HBTC (liver diseases) | EMRs of patients with liver diseases in Hubei provincial hospital of TCM |
| 4 | Five hospitals in Hubei province | HBTC (COVID-19) | EMRs of patients with COVID-19 from five hospitals in Hubei province (namely Hubei Provincial Hospital of Traditional Chinese Medicine, Wuhan Huangpi District Hospital of Traditional Chinese Medicine, Hubei 672 Orthopaedics Hospital of Integrated Chinese & Western Medicine, Wuhan Hospital of Traditional Chinese Medicine, and Wuhan Hospital of Traditional Chinese and Western Medicine) |
| **TCM books and dictionaries** | | | |
| 5 | Differential diagnosis of TCM symptoms | DDTCMS | Differential diagnosis of TCM symptoms is a book published by People's Health Publishing House on March 1, 2000. The author is Yao Naili. This book is an important part of differential diagnosis of traditional Chinese medicine. |
| 6 | Symptomatic study of traditional Chinese medicine | SCM | Symptomatic study of traditional Chinese medicine is a book published by Ancient Books of Traditional Chinese Medicine Publishing House in 2013. The authors are Zhang Qiming and Liu Baoyan. Mainly introduces the definition of symptoms, clinical features of symptoms and classification of symptoms. |
| 7 | Study on standardization of terms of traditional Chinese medicine | SSTTCM | Study on standardization of terms of traditional Chinese medicine collects the results and experience of researcher Zhu Jianping and his team that engaged in the research on the standardization of traditional Chinese medicine terminology for 16 years. It is the basis of standardization, modernization and internationalization of traditional Chinese medicine. |
| 8 | Traditional Chinese Medicine Terms | TCMT | Traditional Chinese Medicine Terms is a book published by the Publishing Department of Science Press in 2005. The author is the Examination Committee of Terms of Traditional Chinese Medicine. |
| 9 | Chinese pharmacopoeia | CP | The Pharmacopoeia of the People's Republic of China (referred to as *Chinese Pharmacopoeia*) is a book published by China Medical Science and Technology Press on June 5, 2015. It was created by the National Pharmacopoeia Committee. Mainly describes the production, extraction, indications of herbs. |
| 10 | Classification and Codes of Tongue Manifestation for Diagnosis in TCM | CTTCM | Classification and Codes of Tongue Manifestation for Diagnosis in TCM (T/CIATCM 010-2019) is an industry standard implemented on May 1, 2019. The standard |

| | | | stipulates the diagnosis name, classification and code of tongue manifestation in TCM. |
|---|---|---|---|
| 11 | Classification and Codes of Pulse Manifestation for Diagnosis in TCM | CPTCM | Classification and Codes of Pulse Manifestation for Diagnosis in TCM (T/CIATCM 011-2019) is an industry standard implemented on May 1, 2019. The standard stipulates the diagnosis name, classification and code of pulse manifestation in TCM. |
| 12 | Ming Yi Hui Cui | MYHC | Ming Yi Hui Cui (Collection of Ancient and Modern Famous Doctors) is a comprehensive medical book edited by Qing Dynasty Luo Mei in 1675. |
| 13 | Lei Zheng Zhi Cai | LZZC | Lei Zheng Zhi Cai is a comprehensive medical book. It was written by Qing Lin Peiqin in 1839. The author makes a detailed analysis of miscellaneous diseases in internal medicine, gynecology, surgery and other diseases according to different etiologies and clinical manifestations. |
| 14 | Lin Zheng Zhi Nan Yi An | LZZN | Lin Zheng Zhi Nan Yi An is a monograph on medical cases written by a famous doctor Ye Tianshi in Qing Dynasty, which records Ye Tianshi's clinical experience. |
| 15 | Yi Fang Kao | YFK | Yi Fang Kao was written by Wu Kun in the Ming Dynasty, recording the names, ingredients, effects, prescriptions, and indications of 780 prescriptions. |
| 16 | Wen Re Jing Wei | WRJW | Wen Re Jing Wei (Warm fabric) is a comprehensive work of traditional Chinese medicine, written by Zhang Zihe of Jin Dynasty. |
| 17 | Shang Han Lun | TFD | Shang Han Lun (Treatise on Febrile Diseases) is a classic work of Chinese medicine written by Zhang Zhongjing in the Eastern Han Dynasty. |
| 18 | Wen Bing Tiao Bian | WBTB | Wen Bing Tiao Bian (Detailed Analysis of Epidemic Warm Diseases) was written by Wu Jutong in the Qing Dynasty (1798) and is an important representative work on Wen Bing (Warm Disease) theory. It is one of the "Four Classics of Traditional Chinese Medicine". |
| 19 | Liu Yin Tiao Bian | LYTB | Liu Yin Tiao Bian (Six Causes Differentiation) is a monograph on exogenous diseases. It was written by Lu Tingzhen in Qing Dynasty in 1868. |
| 20 | Yi Jia Bi Yong | YJBY | Yi Jia Bi Yong is a clinical and comprehensive TCM book compiled by Sun Yingkui in the Ming Dynasty. It was completed in the 32nd year of Jiajing in the Ming Dynasty (1553). |
| 21 | Ben Cao Chong Yuan | BCCY | Ben Cao Chong Yuan (Reverence for the origin of Materia Medica) is a pharmacy monograph annotating "Shen Nong's Materia Medica", written by Zhang Zhicong in the Qing Dynasty (1674). |
| 22 | Ben Cao Dong Quan | BCDQ | Ben Cao Dong Quan (Interpretation of medica hole) is a book on herbal medicine written by Shen Mu in the Qing Dynasty. |
| 23 | Bi Hua Yi Jing | BHYJ | Bi Hua Yi Jing is a clinical classic book of intangible cultural heritage of TCM and is another great medical treasure in China, written by Jiang Hantun in the Qing Dynasty. |
| 24 | Zhu Bing Yuan Hou Lun | ZBYHL | Zhu Bing Yuan Hou Lun (General treatise on causes and manifestations of all diseases) is a book about the etiology and syndrome works of TCM, written by Chao Yuanfang in the Sui Dynasty. |
| 25 | Ru Men Shi Qin | RMSQ | Ru Men Shi Qin is a comprehensive work of TCM, written by Zhang Zihe of Jin Dynasty. |
| **Public available biomedical controlled vocabularies** | | | |
| 26 | Unified Medical Language System | UMLS | Version.2020 |
| 27 | The Human Phenotype Ontology | HPO | Version.2022 |
| 28 | Symptom Ontology | SO | Version.2022 |
| 29 | International Classification of Diseases | ICD-11 | Version.11th |
| 30 | Medical Subject Headings | MeSH | Version.2022 |

## Supplementary Table S2. The amount information on source concepts

| Index | Data Source | Term Number |
|---|---|---|
| **Electronic medical records** | | |
| 1 | Shandong provincial hospital of TCM | 1,817 |
| 2 | Shanxi Provincial Hospital of TCM | 1,130 |
| 3 | Hubei provincial hospital of TCM | 693 |
| 4 | Five hospitals in Hubei province | 323 |
| **Contemporary textbooks and dictionaries** | | |
| 5 | Differential diagnosis of TCM symptoms | 1,208 |
| 6 | Symptomatic study of traditional Chinese medicine | 619 |
| 7 | Study on standardization of terms of traditional Chinese medicine | 419 |
| 8 | Traditional Chinese Medicine Terms | 329 |
| 9 | Chinese pharmacopoeia | 412 |
| 10 | Classification and Codes of Tongue Manifestation for Diagnosis in TCM | 397 |
| 11 | Classification and Codes of Pulse Manifestation for Diagnosis in TCM | 73 |
| **Ancient TCM books** | | |
| 12 | Ming Yi Hui Cui | 305 |
| 13 | Lei Zheng Zhi Cai | 278 |
| 14 | Lin Zheng Zhi Nan Yi An | 226 |
| 15 | Yi Fang Kao | 210 |
| 16 | Wen Re Jing Wei | 148 |
| 17 | Shang Han Lun | 146 |
| 18 | Wen Bing Tiao Bian | 135 |
| 19 | Liu Yin Tiao Bian | 118 |
| 20 | Yi Jia Bi Yong | 98 |
| 21 | Ben Cao Chong Yuan | 72 |
| 22 | Ben Cao Dong Quan | 68 |
| 23 | Bi Hua Yi Jing | 48 |
| 24 | Zhu Bing Yuan Hou Lun | 41 |
| 25 | Ru Men Shi Qin | 29 |
| **Public available biomedical controlled vocabularies** | | |
| 26 | Unified Medical Language System | 9,431 (2,112 concepts) |
| 27 | The Human Phenotype Ontology | 353 (139 concepts) |
| 28 | Symptom Ontology | 637 (476 concepts) |
| 29 | International Classification of Diseases (Version.11) | 220 |
| 30 | Medical Subject Headings | 2,129 (245 concepts) |

**Supplementary Figure S1. Disease distribution in Chinese electronic medical records collected by ISPO.**

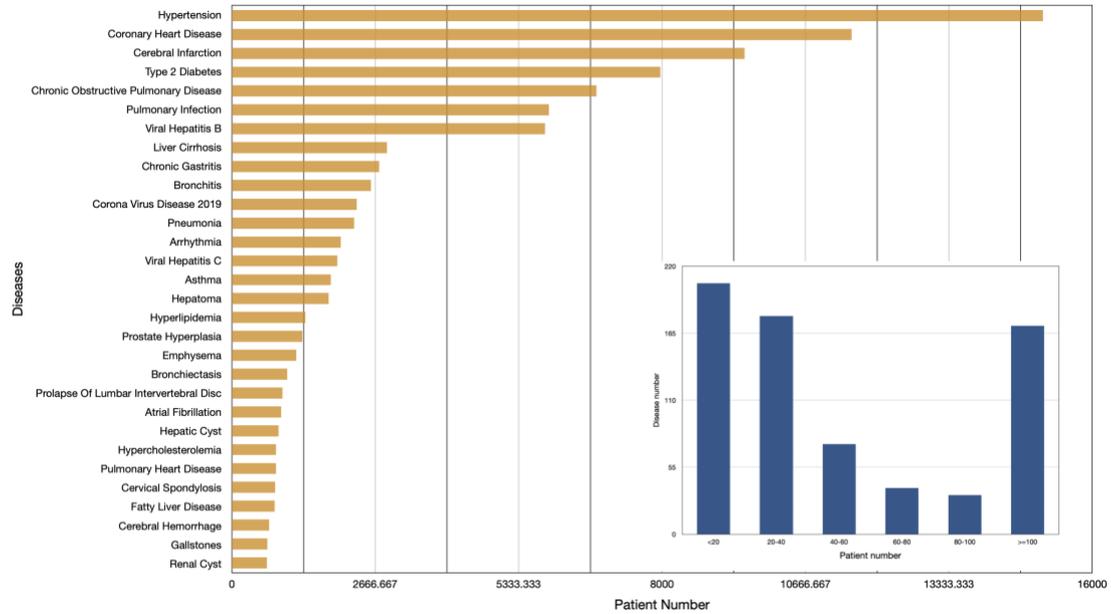

We collected Chinese EMRs of 78,696 inpatient cases from 7 highest-level TCM hospitals. In our dataset, there is a long-tail distribution of disease entities, which is a common phenomenon in clinical setting. There are over 160 diseases with more than 100 cases, and nearly 200 diseases with fewer than 20 cases. We have shown the distribution of 30 diseases with over 500 cases, such as chronic lung disease, heart disease, stroke, cancer, hypertension and kidney disease.

**Supplementary Figure S2. The distribution characteristics of system category for the different sources**

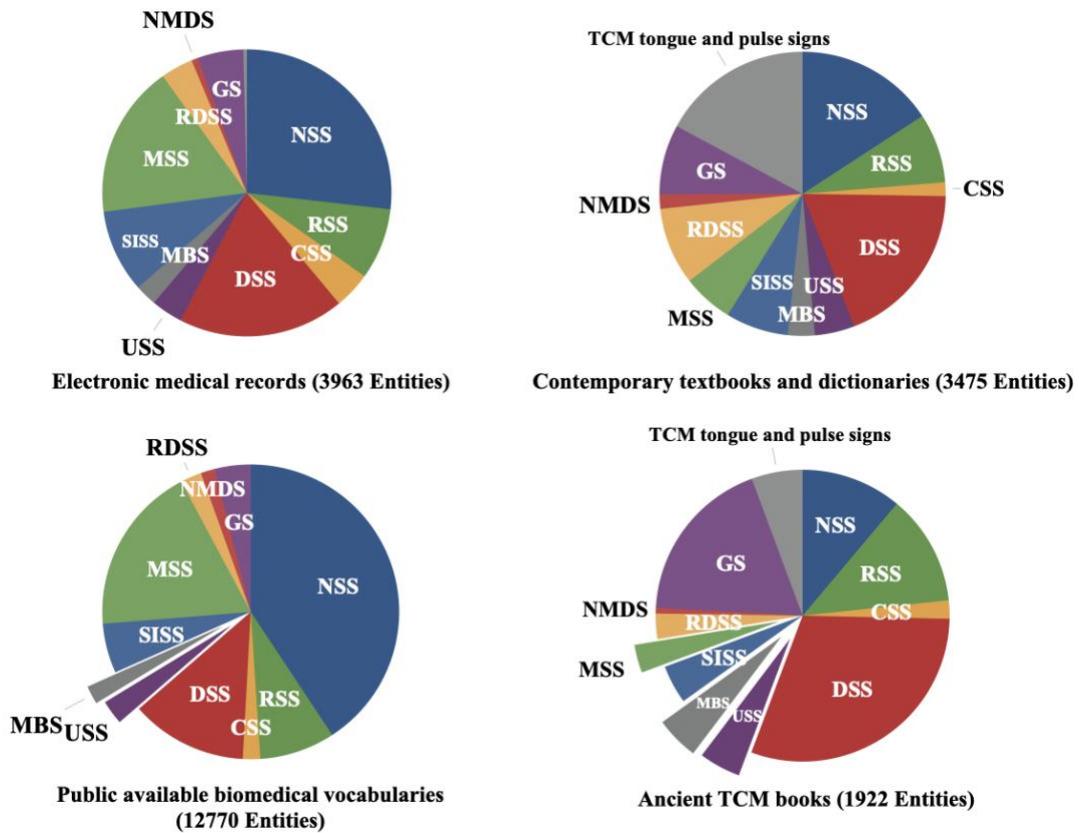

We compared the characteristics of system categories for concepts from different sources of knowledge by calculating proportions. Abbreviation: Nervous system symptoms (NSS), Respiratory system symptoms (RSS), Circulatory system symptoms (CSS), Digestive system symptoms (DSS), Urinary system symptoms (USS), Mental and behavioral symptoms (MBS), Skin and integumentary system symptoms (SISS), Musculoskeletal system symptoms (MSS), Reproductive system symptom (RDSS), Nutrition, metabolism, and development symptoms (NMDS), General symptoms (GS).